\pdfoutput=1
\documentclass[11pt]{article}

\usepackage[]{EMNLP2022}

\usepackage{times}
\usepackage{latexsym}

\usepackage[T1]{fontenc}
\usepackage[utf8]{inputenc}

\usepackage{microtype}

\usepackage{inconsolata}

\usepackage{graphicx}
\usepackage{booktabs}
\usepackage{amsfonts}

\title{Biomedical NER for the Enterprise with Distillated BERN2 and the Kazu Framework}

\author{
  Wonjin Yoon \textsuperscript{1}\textsuperscript{$\dagger$} \\\AnDD
  Richard Jackson \textsuperscript{2}\textsuperscript{$\dagger$} \\\AnD
  Elliot Ford \textsuperscript{2} \\\AnD
  Vladimir Poroshin \textsuperscript{2} \\\AnDD
  Jaewoo Kang \textsuperscript{1,3} \\
  \end{tabular}\hss\egroup \hfil\hfil\egroup
           \hbox to \linewidth\bgroup\large \hfil\hfil
             \hbox to 0pt\bgroup\hss \begin{tabular}[t]{c}
  \textsuperscript{1}Korea University, Seoul, South Korea\\
  \textsuperscript{2}AstraZeneca, Cambridge, United Kingdom \\
  \textsuperscript{3}AIGEN Sciences, Seoul, South Korea \\
  \textsuperscript{$\dagger$} Equal contribution \\
  \texttt{\{wjyoon, kangj\}@korea.ac.kr}\\
  \texttt{\{richard.jackson4, elliot.ford, vladimir.poroshin\}@astrazeneca.com}
}

\makeatletter
\newcommand{\printfnsymbol}[1]{
  \textsuperscript{\@fnsymbol{#1}}
}

\makeatother

\begin{document}
\maketitle
\begin{abstract}

In order to assist the drug discovery/development process, pharmaceutical companies often apply biomedical NER and linking techniques over internal and public corpora. Decades of study of the field of BioNLP has produced a plethora of algorithms, systems and datasets. However, our experience has been that no single open source system meets all the requirements of a modern pharmaceutical company. In this work, we describe these requirements according to our experience of the industry, and present Kazu, a highly extensible, scalable open source framework designed to support BioNLP for the pharmaceutical sector. Kazu is a built around a computationally efficient version of the BERN2 NER model (TinyBERN2), and subsequently wraps several other BioNLP technologies into one coherent system.

\end{abstract}

\section{Introduction}

One of the promises of the applications of A.I. within the pharmaceutical sector is to empower the search for new drugs, and quicken their development into safe, effective medicines. Within the field of NLP, this commonly involves the application of named entity recognition (NER, which is the task of finding entities from a document) and entity linking (EL, also known as grounding, or normalisation), and other techniques to internal and external documents. Documents enhanced with such metadata have a wide variety of use cases, such as improving the performance of enterprise search systems, phase 4 monitoring of adverse events or as a precursor to relationship extraction (for instance in biomedical knowledge graph construction \cite{geleta2021biological}).

Our experience has been that high quality NER remains at the core of many typical NLP use cases within the pharmaceutical industry, and therefore is the prominent focus of our work. BioNER as a field is notable for it's technical complexity and chronic shortage of sufficiently sized training/test datasets, relative to general domain corpora. Although recent advances have produced excellent results on benchmark datasets, recent work \cite{kim2022your} has also suggested that such approaches may be overfit, and may not necessarily generalise sufficiently to meet the needs of a production system. 

Similarly, the tendency of academic products to focus on minimising the error rate over a given benchmark ignores practical issues of productionisation, such as the computational complexity of an algorithm, ecological impact and the maintenance of a coherent codebase. While a low overall error rate is undoubtedly important for any enterprise A.I. system, world class performance often comes at the expense of speed. Therefore, striking a balance between an acceptable error rate and other performance metrics is central to user acceptance. We posit that `near' (rather than `absolute') state of the art is sufficient for most use cases.

While several commercial solutions are available to address this requirement, we suggest that a freely available open source solution to deal with the intricacies of this area has not been forthcoming. In this piece, we describe the practical challenges and requirements of enterprise BioNLP analytics. We present our TinyBERN2 biomedical NER model, which utilises weak supervision to address generalisability issues,  and our associated Kazu framework by which we deploy it for enterprise applications within a large pharmaceutical company. 
The Kazu framework and models (including TinyBERN2 and distillated PubMedBERT \cite{gu2021domain}) are open-sourced \footnote{\url{https://github.com/AstraZeneca/KAZU}}.

\section{Challenges of BioNLP in the Pharmaceutical Sector and the Kazu Framework}

The priorities of academic research in NLP often do not focus on the various practical elements of productionising algorithms within the context of a corporate environment. Nevertheless, this is one of the domains where their outputs can deliver value - via extending existing systems and enabling new projects. The challenges of managing A.I. systems in such environments are acknowledged by the emergence of the field of MLOps, and we repeat some of the most salient aspects relevant to BioNLP here.

\subsection{Language/technology agnostic and scalability} 

The majority of algorithms for BioNLP are typically written in JVM languages or Python, each of which may have dependency conflicts with other algorithms within the overall Kazu pipeline. Here, we utilise the scalable Ray framework \cite{ray-222605} which allows different processes to run with distinct Python virtual environments/JVM classpaths, substantially reducing the chance of a conflict.

\subsection{Flexibility of datasource ingestion} 

The biomedical domain is awash with ontologies and knowledgebases, representing various attempts to standardise and model biological concepts. These typically form the basis of EL targets and/or dictionary based NER vocabularies. Thus, we have built a parsing system to allow any data source to be converted into a vocabulary, suitable for curated dictionary based entity matching and/or entity linking.

\subsection{Robustness of data model} 

The biomedical literature is known for the over-representation of certain linguistic phenomena, such as multi section documents/abstracts, nested entities \cite{Alex2007RecognisingNN} and non-contiguous entities \cite{Lever2020ExtendingTF}. We note that the data models of many popular NLP frameworks don't contain native support for these concepts, and have thus built these into the standard Kazu dataclasses.

\subsection{Extensibility of pipeline design} The current pace of NLP development is extremely rapid. We present Kazu with implementations of several algorithms that we have chosen based on our preference at the time of writing. However, we recognise that any or all of these are likely to be super-ceded in the short to medium term. Therefore, we have designed Kazu in a modular pipeline fashion, wherein new algorithms can be introduced relatively easily.

\subsection{Stability in execution} Executing NLP algorithms over large corpora of text is notoriously unreliable, due to the difficulties in building systems that are able to cope with highly arbitrary input. For instance, certain strings of text may cause NER processes to crash due to high memory usage. Systems that are able to identify problematic text, and either a) avoid processing exceptions or b) recover from such situations are helpful in production environments. To deal with these scenarios, we leverage several techniques such as process memory monitoring/automatic worker restarting, which in turn allows the processing of millions of documents with relative ease.

\section{Methods}
\subsection{Model Architecture}

Historically, BioNER approaches have utilised `sequence tagging' style tasks \cite{yoon2019collabonet, lee2020biobert}, wherein a string of text is tokenized, and a model assigns each token a label according to the popular Begin, Inside, Outside schema. 
However, this single-label classification approach can not properly predict \textit{nested entities} \cite{katiyar-cardie-2018-nested}, where the spans of multiple types of entities are overlapped, without additional processing or methods. 

To mitigate the problem from nested entity, we applied multi-label label prediction for each token. The method is straight-forward and neither require complicated training strategy nor additional parameters that leads to significant speed reduction in inference.

Our model is composed of BERT layers and a dense output layer. For the case where the number of entity classes is $k$ and using the B-<entity class>, I-<entity class>, and O tag schema  \cite{Ramshaw1999}, the output layer $o$ is defined as follows:
\begin{equation}
    o = Sigmoid(h  W  + b)
\end{equation}
where $h \in \mathbb{R}^{d} $ is the output of the final layer of BERT layers for a token $W \in \mathbb{R }^{d*(2*k+1)}$, and $b \in \mathbb{R}^{2*k+1}$. The output layer will produce a vector (with $2*k+1$ elements) of probability for each tokens.

The training objective is to reduce loss between the output of the model (vector of size $2*k+1$) and the annotation vector. Each element of the annotation vector represents whether the token is a part of the corresponding entity class. 
Note that the elements are independent and an annotation vector can label multiple entity classes.
The element can be a binary value (hard-label) or a probability value (soft-label) of an off-the-shelf model. We used a standard Binary Cross Entropy loss (as implemented in the pytorch library) as our loss function \footnote{\url{https://pytorch.org/docs/1.12/generated/torch.nn.BCELoss.html}}.

\begin{figure*}[h] 
\includegraphics[width=\textwidth]{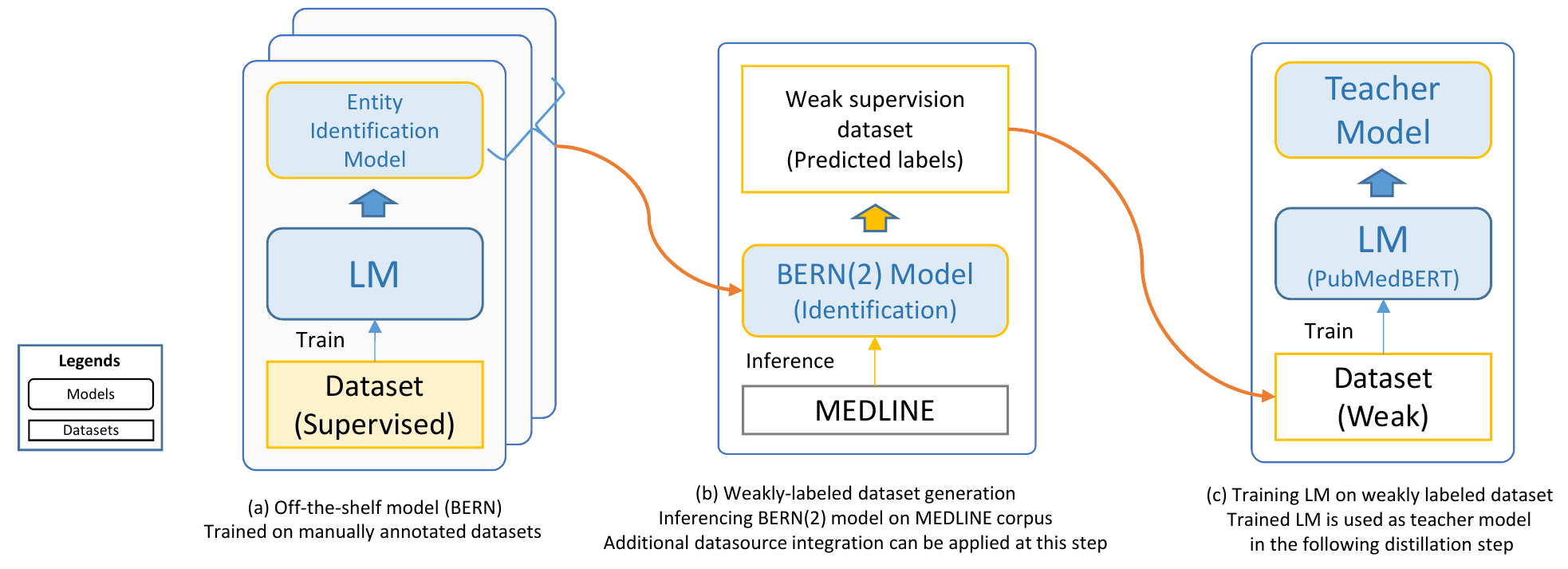}
\caption{Building a weakly supervised dataset and subsequent training our weakly supervised BERN2 (WS-BERN2) model. We used BERN2 \cite{sung2022bern2} to generate weakly-labeled datasets. Alternatively, off-the-shelf models / existing models can be used. Additional datasource integration can be applied at the step (b). Generated weakly-labeled datasets can be used to train a full-sized LM (WS-BERN2, or teacher model in the step (c)) or to train a tiny-sized LM (distillated LM) directly.} \label{fig:datageneration}
\end{figure*}

\subsection{Weakly supervised learning}\label{sec:model-sub-weak-super}

To address the aforementioned generalisability concerns, we adapt a weakly supervised learning strategy \cite{Ratner2018SnorkelMW}.
Figure~\ref{fig:datageneration} shows an example of our training dataset generation. 
Off-the-shelf models or existing models can be used to generate weakly-labeled datasets. In our case, we utilized predictions of BERN2 \citep{sung2022bern2} on PubMed articles, that is available on the official web-page \footnote{\url{http://bern2.korea.ac.kr/}} (downloaded Feb 7, 2022 version v1.0).  
Statistics of the downloaded dataset are shown in Table~\ref{tab:bern-dump-stats}. We pre-processed about 25 Million MEDLINE abstracts available in PubMed website \footnote{\url{https://www.nlm.nih.gov/databases/download/pubmed_medline.html}}.
In order to reduce ecological footprint during experiments, we used about 10\% of BERN2-labeled dataset to make our weakly labeled dataset. Articles in test set of benchmark datasets are filtered out from our weakly-labeled training dataset by PMIDs to prevent any downstream models from directly learning/memorizing of BERN2 predictions.

Some of the obstacles in training with weakly supervised learning are noises and biases added during the labeling  \cite{jiang-etal-2021-named}. In the Section ~\ref{sec:dis-sub-soft}, we explore both soft-labels (i.e. training with the confidence values of the supervising model) and hard-labels (i.e. training with the categorical labels produced by the supervising model).

\subsection{Distillation}

Distillation is a key tool to address the scalability requirement as it enables to make computationally efficient models while retaining most of the F1 performance. During distillation, both the hidden size and the number of layers are reduced. The former reduces the parameter size, resulting a smaller memory usage, and eventually facilitates inferencing with much larger batchsize. The latter not only reduces memory usage, but also decreases the CPU/GPU time spent for a single example to be processed.

Following the work of \citet{jiao-etal-2020-tinybert}, we applied two-stage approach of distillation. 
In the first stage, we distillate a biomedical domain specific transformer language model to build a  task-independent tiny language model (LM). 
In the second stage, the distilled LM can be trained for a task-specific dataset (such as an NER task) directly or alternatively, a full-sized transformer model trained on the specific NER task is used as a teacher model, which in turn tunes a task-specific distilled LM.

In our experiment, the teacher model for the first stage is PubMedBERT \cite{gu2021domain}. For the second stage, we first train a full sized BERN2 \citep{sung2022bern2} on our weakly labeled dataset, which is then used as a teacher model for our final, distilled NER model (TinyBERN2).

\section{Experiments and Results}

\begin{table}[]
\centering
\resizebox{\columnwidth}{!}{%
\begin{tabular}{@{}lrr@{}}
\toprule
Type & BERN2-labeled & Our weakly-labeled \\\midrule
Abstracts           &  25,726,681    &    \\
Sentences           &  157,267,033   &  16,712,485  \\
Words (tokens)      &  3,646,395,389 &  438,686,717  \\
Words per sentences &  23.18         &  26.24  \\ \bottomrule
\end{tabular}
}
\caption{Statistics of BERN2-labeled dataset and our weakly-labeled dataset. About 10\% of the BERN2-labeled dataset is used as the weakly-labeled dataset for the training step of TinyBERN2. \textit{Words} denotes the tokens delimited by spaces or special characters in the sentence.}
\label{tab:bern-dump-stats}
\end{table}

\subsection{Benchmark Datasets}

For evaluation of our TinyBERN2 model, we chose 8 benchmark datasets of 6 entity classes: Gene/Protein, Disease, Chemical, Species, Cell line, and Cell type \cite{dougan2014ncbi,li2016biocreative,krallinger2015chemdner,smith2008overview,kim2004introduction,gerner2010linnaeus,neves2013preliminary, Kaewphan2016CellLN}.
While we were designing the experiment, we wanted to examine the generalizability (i.e. ability to predict \textit{unseen entities} that are not in the training dataset \citep{kim2022your}, which is essential for real-world use cases. To this end, we evaluate our model on several datasets that were not used to train our teacher model (BERN2). These are marked with $\dagger$ in the main experiment table (Table~\ref{tab:performance}).

For the benchmark datasets, we used \textit{MTL-Bioinformatics-2016} GitHub repository \footnote{\url{https://github.com/cambridgeltl/MTL-Bioinformatics-2016}} \cite{crichton2017neural} with an additional processing step that ensures all special characters are consistently used as token delimiters. All benchmark datasets were pre-processed to have the same format, where a line contains one token and a corresponding label tag as in CoNLL-X format \cite{buchholz-marsi-2006-conll}.

\subsection{Results}
Table~\ref{tab:performance} shows our experimental results on benchmark datasets. We compared the performance of the off-the-shelf model (BERN2), weakly-supervised model (WS-BERN2), and distilled weakly-supervised BERN2 model (TinyBERN2) in terms of precision/recall/F1 and computational costs.

\begin{table*}[t]
\centering
\resizebox{\textwidth}{!}{%
\begin{tabular}{@{}llcccccc@{}}
\toprule
                                     &  &     & BERN2 (*)                   &         WS-BERN2             & TinyBERN2             \\ \midrule
       Group& Benchmark              &  & P / R / F1                  &         P / R / F1           & P / R / F1                   \\ \midrule
Gene        & BC2GM $^\dagger$       &  & 82.47\% / 80.77\% / 81.61\% & 82.52\% / 82.55\% / 82.54\%  & 80.80\% / 79.37\% / 80.08\%  \\
            & JNLPBA                 &  & 67.23\% / 72.37\% / 69.70\% & 65.88\% / 70.14\% / 67.95\%  & 64.72\% / 69.08\% / 66.83\%   \\
            & CellFinder             &  & 63.28\% / 52.89\% / 57.62\% & 70.25\% / 80.40\% / 74.98\%  & 63.67\% / 75.71\% / 69.17\%  \\ \midrule
Disease     & NCBI-disease $^\dagger$&  & 86.33\% / 80.94\% / 83.55\% & 87.41\% / 88.23\% / 87.82\%  & 84.95\% / 82.92\% / 83.92\%  \\
            & BC5CDR-Disease         &  & 77.14\% / 64.76\% / 70.41\% & 78.26\% / 67.36\% / 72.40\%  & 76.91\% / 66.25\% / 71.18\%  \\ \midrule
Chemical    & BC4CHEMD $^\dagger$    &  & 93.66\% / 92.06\% / 92.86\% & 91.93\% / 91.44\% / 91.69\%  & 90.02\% / 88.70\% / 89.36\%  \\
            & BC5CDR-Chemical        &  & 94.41\% / 86.57\% / 90.32\% & 94.55\% / 87.59\% / 90.94\%  & 94.52\% / 86.55\% / 90.36\%  \\ \midrule
Cell line   & JNLPBA $^\dagger$      &  & 50.00\% / 76.10\% / 60.35\% & 33.43\% / 71.80\% / 45.62\%  & 34.85\% / 68.80\% / 46.27\%  \\
            & CellFinder             &  &  8.52\% / 36.07\% / 13.78\% & 10.08\% / 45.02\% / 16.48\%  &  3.61\% / 12.37\% /  5.59\%  \\
            & GELLUS                 &  &  8.61\% / 25.70\% / 12.90\% & 9.01\% / 24.02\% / 13.11\%   &  7.54\% / 18.99\% / 10.79\%  \\ \midrule
Cell type   & JNLPBA $^\dagger$      &  & 60.49\% / 67.96\% / 64.01\% & 33.43\% / 71.80\% / 65.33\%  & 66.41\% / 64.06\% / 65.22\%  \\
            & CellFinder             &  & 51.12\% / 32.19\% / 39.50\% & 48.86\% / 31.81\% / 38.53\%  & 52.33\% / 25.06\% / 33.89\%  \\ \midrule
Species     & LINNAEUS $^\dagger$    &  & 89.13\% / 91.56\% / 90.33\% & 80.82\% / 44.01\% / 56.99\%  & 81.86\% / 42.05\% / 55.56\%  \\
            & CellFinder             &  & 33.38\% / 75.08\% / 46.21\% & 38.03\% / 52.96\% / 44.27\%  & 46.02\% / 50.47\% / 48.14\%  \\ \midrule \midrule
Model complexity  & Backbone model        &     & Bio-LM \cite{lewis-etal-2020-pretrained} & PubMedBERT \cite{gu2021domain} & Distilled model \\
                  & \# Parameters         &     &         365M                &       109M                   &     14M    \\
                  & \# Layers             &     &         24                  &             12               &          4                    \\ \midrule  \midrule
Throughput (CPU)  & Speed (s/steps)      &     &       0.37 sec/steps       &        0.043 sec/steps&      0.017 sec/steps  \\
 (batch size = 1) & Throughput (samples/s)&     &      2.66  samples/s &       22.93  samples/s &      57.43   samples/s \\\midrule
 Throughput (CPU)  & Speed (s/steps)      &     &       N/A              &        1.204 sec/steps&      0.119 sec/steps  \\
 (batch size = 32) & Throughput (samples/s)&     &      N/A              &       26.56  samples/s &      267.40   samples/s \\\midrule
Throughput (GPU)  & Speed (s/steps)       &     &      0.037 sec/steps           &        0.012 sec/steps            &           0.006 sec/steps \\
 (batch size = 1) & Throughput (samples/s)&     &           26.33 samples/s      &          83.66   samples/s         &               160.86  samples/s \\\midrule
Throughput (GPU)  & Speed (s/steps)      &     &           N/A             &          0.086 sec/steps          &          0.023 sec/steps          \\
 (batch size = 32)& Throughput (samples/s)&     &         N/A             &          367.89    samples/s          &          1354.08  samples/s          \\ \bottomrule
\end{tabular}%
}
\caption{Performance of our TinyBERN2 model and BERN2 model. 
Benchmark datasets that are used to train BERN2 is marked with $\dagger$.
\textit{WS-BERN2} denotes our model trained on weakly-supervised dataset.
\textit{\# Layers} denotes the number of transformers layer in the backbone model, excluding embedding layer and the output layer.
*: Performance for the BERN2 may vary with the BERN2 paper \cite{sung2022bern2} as we applied different tokenization schema, and application overheads for throughput.
}
\label{tab:performance}
\end{table*}

\subsubsection{Evaluation metrics (accuracy)}
We measured entity level Precision, Recall, and F1-score using \textbf{SeqEval} library \footnote{\url{https://huggingface.co/spaces/evaluate-metric/seqeval}} \cite{seqeval}. 
Some datasets contain multiple entity classes. As our model supports multi-label output for each token, we first save model predictions for all types and collect each type separately using the saved output and evaluate entity classes using the collected output. 
Formally, assume that a model can predict $k$ entity classes and an example has $n$ tokens. If we use BIO-tagging schema the number of labels are $2*k+1$ including "O" label. The output of the model can be denoted as a matrix $M \in \mathbb{R}^{(2*k+1) \times n}$. For evaluating $i$-th ($i \in \{1 \cdots k\}$) entity class, we use two rows that marks the given entity class and a row for "O" in $M$ each of them is a vector of length $n$. These three vectors are merged and form a prediction list, which is used along with the gold standard labels for evaluating an entity class of a benchmark dataset.

In our experiments, F1-scores of our weakly supervised model (WS-BERN2) were analogous to BERN2 model for entity classes where more training data were available (Gene, Disease, Chemical). Our weakly supervised model showed better performance in cross-dataset evaluation (i.e. evaluation on datasets that are not used to train BERN2), which support our assumption on generalisability. As expected, our distilled model, TinyBERN2 showed a lower F1 score across most datasets compared to WS-BERN2, although this was marginal in many cases.

\subsubsection{Evaluation metrics (Computational costs)}

For evaluation of the models in respect of throughput and speed, we used 26,365 sentences from BC4CHEMD test dataset. One sentence forms a test sample. 
\textit{Speed} in Table~\ref{tab:performance} is a measurement for seconds spent for an evaluation step: we divided the number of steps by time spent to process the dataset using mini batch size of 1.  The models were loaded in the memory before the experiment. 

For BERN2 model (the off-the-shelf model), we installed BERN2 in a local machine to reduce network overheads. The model codes were modified to run without normalization and rule-based post-processing features, with help of the BERN2 authors. For evaluating the memory usage of BERN2, we only report memory usage of batch\_size = 1 setting, as BERN2 doesn't currently support making batch predictions \footnote{\url{https://github.com/dmis-lab/BERN2/issues/10}}. For the same reason, we omit throughput results for BERN2.

We used a bare-metal server (Ubuntu Server 16.04.3 LTS) with single NVIDIA TITAN Xp (12GB) GPU and Xeon(R) E5-2630 v4 @ 2.2GHz (10 Core / 20 Threads) CPU for the experiments.

\section{Discussions}

\subsection{Effect of training by soft-labeling}\label{sec:dis-sub-soft}
In Section~\ref{sec:model-sub-weak-super} we suggested that the traditional binary labeling, or hard-labeling, can hinder the weakly-labeled training by adding biases. To examine this hypothesis, we compared the performance of models trained using the hard-label dataset and the soft-label dataset.

Table~\ref{tab:performance-tagging} shows the performance of WS-BERN2 models trained on BIO-tagging with hard-labels and soft-labels (for the comparison between IO-tagging and BIO-tagging, see Section~\ref{sec:dis-sub-tagging-sch}). In macro average score, WS-BERN2 model trained with soft-label outperformed model trained with hard-label by 0.52\%. 
An interesting observation is that the model benefited from soft-labelling for entity classes with fewer training instances in the weakly-labeled dataset. Here, the macro average improvement was 0.89\% in 7 benchmark datasets across the species, cell line and cell type entity classes. Marginal improvements were observed for entity classes where more training data were available (Gene, Disease, Chemical), with a macro average improvement of 0.14\% in 7 benchmark datasets.

\begin{table}[]
\centering
\resizebox{\columnwidth}{!}{%
\begin{tabular}{@{}llcccccc@{}}
\toprule
            &                         &BIO-Hard  &BIO-Soft &IO-Soft       \\ \midrule
Entity Group& Benchmark               & F1       & F1      & F1          \\ \midrule
Gene        & BC2GM $^\dagger$        &82.78\%   & 82.54\% & 83.76\%   \\
            & JNLPBA                  &67.93\%   & 67.95\% & 68.82\%   \\
            & CellFinder              &73.71\%   & 74.98\% & 75.36\%   \\ \midrule
Disease     & NCBI-disease $^\dagger$ &87.38\%   & 87.82\% & 87.82\%   \\
            & BC5CDR-Disease          &72.67\%   & 72.40\% & 72.85\%   \\ \midrule
Chemical    & BC4CHEMD $^\dagger$     &91.81\%   & 91.69\% & 92.44\%   \\
            & BC5CDR-Chemical         &91.05\%   & 90.94\% & 91.26\%   \\ \midrule
Cell line   & JNLPBA $^\dagger$       &44.70\%   & 45.62\% & 48.27\%   \\
            & CellFinder              &15.66\%   & 16.48\% & 14.73\%   \\
            & GELLUS                  &12.46\%   & 13.11\% & 12.11\%   \\ \midrule
Cell type   & JNLPBA $^\dagger$       &64.45\%   & 65.33\% & 66.60\%   \\
            & CellFinder              &39.27\%   & 38.53\% & 38.83\%   \\ \midrule
Species     & LINNAEUS $^\dagger$     &56.70\%   & 56.99\% & 57.23\%   \\
            & CellFinder              &40.86\%   & 44.27\% & 43.33\%   \\ \midrule \midrule
            & Macro Average           &60.10\%   & 60.62\% & 60.96\%   \\ \bottomrule
\end{tabular}%
}
\caption{Performance of our weakly-supervised BERN2 models by different labeling schema. 
Benchmark datasets that are used to train BERN2 is marked with $\dagger$.
}
\label{tab:performance-tagging}
\end{table}

\subsection{Tagging Schema}\label{sec:dis-sub-tagging-sch}

Previous works on Biomedical NER tasks have preferred to use the BIO-tagging schema over IO-tagging (i.e. only tag with Inside or Outside), presumably because BIO-tagging can delimit entity spans completely (i.e. IO-tagging cannot delimit two consecutive entities). However, we revisit this convention for the NER task where the model is a transformer based language model and the input text are from the formal scientific literature. 

Transformer based language models, such as BERT and BioBERT, use trained tokenizers, such as BPE-tokenizers \cite{sennrich-etal-2016-neural}, and do not remove stopwords or special characters. Instead those tokenizers make stopwords or special characters an independent token.

We hypothesized that in scientific literature, the writer of the text tend to express themselves in a way that causes entities to be wrapped with non-entities, stopwords (such as \textit{and}, \textit{or}) or punctuation characters (such as "\textit{,}", "\textit{.}", "\textit{'}", and "\textit{"}").  

From this perspective, we conducted a preliminary study using BERN2 predictions on 106,921 sentences from randomly sampled articles. Across 102,569 entities, 99,814 entities (97.3\%) do not have adjacent entities, in the sense that an entity span starts right after an entity span of the same class ends - only 466 entities were found to have adjacent entities that cannot be delimited with the IO-tagging schema.

\begin{figure}[h] 
\includegraphics[width=\columnwidth]{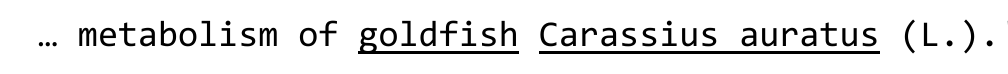}
\caption{An example of adjacent entities (\textit{goldfish} and \textit{Carassius auratus}).} \label{fig:adjacent}
\end{figure}

Based on this observation, we conducted an experiment to evaluate the usability of IO-tagging. 
Performances of WS-BERN2 models trained on IO-soft and BIO-soft are denoted in Table~\ref{tab:performance-tagging}. Based on the macro average score, the model using IO-tagging outperformed the model with BIO-schema. For the performance on entity classes across the benchmark datasets, 11 scores output 14 were improved by using IO-tagging. Using IO-tagging was also beneficial for the aspect of the computational cost required for the training convergence (i.e. fully trained), as the model is less complex. 

However, we do not recommend the IO-tagging model for enterprise use cases, as the input samples cannot be guaranteed to be restricted in the scientific/academic writing with complete punctuation. Absence of statistics for manually-labeled dataset remains as limitation of this auxiliary study and is a topic for further research. 

\subsection{Enterprise usage of Kazu}

We have integrated our TinyBERN2 model into our new Kazu framework, alongside other components for abbreviation expansion \cite{neumann-etal-2019-scispacy}, entity linking \cite{DBLP:journals/corr/abs-2010-11784} and additional novel algorithms outside the scope of this paper (full details of the other Kazu components can be found at \footnote{\url{https://github.com/AstraZeneca/KAZU}}).

Kazu is deployed at AstraZeneca enterprise wide, and is already in use as a core component for projects such as biological knowledge graph (BIKG) construction \cite{geleta2021biological} and clinical trial design via enabling the structured search of clinical studies. An execution over PubMed (abstracts) and PubMed Central (full text documents) has extracted the following (uniquely mapped) references:  22 532  diseases, 19 884 genes, 18 715 drugs, 6469 anatomy references, 5 372 cell line references, and 53 cell type references. Additional deployments of Kazu for other internal projects are planned in the near future.

\section*{Limitations}

One of our objectives in this work was to establish that the TinyBERN2 model is highly performant, both in terms of accuracy metrics and computational efficiency. However, we were limited to testing this on just a single type of CPU and GPU. Since hardware varies dramatically, it is difficult to predict the precise throughput gains for any single setup. In addition, BERN2 incorporates several elements of pre/post processing, that it wasn't possible to completely disable during throughput testing, which will impact the reported throughput and accuracy results to a small degree. Nevertheless, we expect our conclusions to be broadly consistent, given the massive reduction in parameters/layers of the TinyBERN2 model vs the original BERN2.

Notably, both the full BERN2 and our weakly supervised/distilled versions struggled to achieve high F1 on the cell line/type classes. This may be due to inconsistencies in the annotation schema of the cell line/type datasets. However, we also suspect that this is due to the fact our model does not make use of dictionary features. This is consistent with the observations of Kaewphan et al \cite{Kaewphan2016CellLN}, in that ML models without dictionary support tend to perform poorly on this entity class. Further work will seek to supplement our approach to weak supervision with such dictionary features for entity classes that are likely to benefit.

Regarding the sizing of our TinyBERN2 model, we followed the recommendations of the original TinyBERT paper, and did not attempt a hyperparameter search to find the optimal trade off between throughput and performance degradation. Further work should explore this aspect.

\section*{Ethics Statement}
This work complies with the ACL Ethics Policy.

\section*{Acknowledgements}

We would like to express our gratitude to Antoine Lain (University of Edinburgh) for helping authors to collect and unify the format of benchmark datasets and Mujeen Sung and Minbyul Jeong for providing NER predictions and information for the inference speed experiments.
This work is partially funded by National Research Foundation of Korea [NRF-2020R1A2C3010638], the Korea Health Industry Development Institute (KHIDI) [HR20C0021] and ICT Creative Consilience program [IITP-2022-2020-0-01819] funded by Government of Republic of Korea. We would also like to thank Rolando Fernandez for his development work on Kazu.

\bibliography{anthology,custom}

\begin{thebibliography}{29}
\expandafter\ifx\csname natexlab\endcsname\relax\def\natexlab#1{#1}\fi

\bibitem[{Alex et~al.(2007)Alex, Haddow, and Grover}]{Alex2007RecognisingNN}
Beatrice Alex, Barry Haddow, and Claire Grover. 2007.
\newblock Recognising nested named entities in biomedical text.
\newblock In \emph{BioNLP@ACL}.

\bibitem[{Buchholz and Marsi(2006)}]{buchholz-marsi-2006-conll}
Sabine Buchholz and Erwin Marsi. 2006.
\newblock \href {https://aclanthology.org/W06-2920} {{C}o{NLL}-{X} shared task
  on multilingual dependency parsing}.
\newblock In \emph{Proceedings of the Tenth Conference on Computational Natural
  Language Learning ({C}o{NLL}-X)}, pages 149--164, New York City. Association
  for Computational Linguistics.

\bibitem[{Crichton et~al.(2017)Crichton, Pyysalo, Chiu, and
  Korhonen}]{crichton2017neural}
Gamal Crichton, Sampo Pyysalo, Billy Chiu, and Anna Korhonen. 2017.
\newblock A neural network multi-task learning approach to biomedical named
  entity recognition.
\newblock \emph{BMC bioinformatics}, 18(1):1--14.

\bibitem[{Do{\u{g}}an et~al.(2014)Do{\u{g}}an, Leaman, and Lu}]{dougan2014ncbi}
Rezarta~Islamaj Do{\u{g}}an, Robert Leaman, and Zhiyong Lu. 2014.
\newblock Ncbi disease corpus: a resource for disease name recognition and
  concept normalization.
\newblock \emph{Journal of biomedical informatics}, 47:1--10.

\bibitem[{Geleta et~al.(2021)Geleta, Nikolov, Edwards, Gogleva, Jackson,
  Jansson, Lamov, Nilsson, Pettersson, Poroshin et~al.}]{geleta2021biological}
David Geleta, Andriy Nikolov, Gavin Edwards, Anna Gogleva, Richard Jackson,
  Erik Jansson, Andrej Lamov, Sebastian Nilsson, Marina Pettersson, Vladimir
  Poroshin, et~al. 2021.
\newblock Biological insights knowledge graph: an integrated knowledge graph to
  support drug development.
\newblock \emph{Biorxiv}.

\bibitem[{Gerner et~al.(2010)Gerner, Nenadic, and Bergman}]{gerner2010linnaeus}
Martin Gerner, Goran Nenadic, and Casey~M Bergman. 2010.
\newblock Linnaeus: a species name identification system for biomedical
  literature.
\newblock \emph{BMC bioinformatics}, 11(1):85.

\bibitem[{Gu et~al.(2021)Gu, Tinn, Cheng, Lucas, Usuyama, Liu, Naumann, Gao,
  and Poon}]{gu2021domain}
Yu~Gu, Robert Tinn, Hao Cheng, Michael Lucas, Naoto Usuyama, Xiaodong Liu,
  Tristan Naumann, Jianfeng Gao, and Hoifung Poon. 2021.
\newblock Domain-specific language model pretraining for biomedical natural
  language processing.
\newblock \emph{ACM Transactions on Computing for Healthcare (HEALTH)},
  3(1):1--23.

\bibitem[{Jiang et~al.(2021)Jiang, Zhang, Cao, Yin, and
  Zhao}]{jiang-etal-2021-named}
Haoming Jiang, Danqing Zhang, Tianyu Cao, Bing Yin, and Tuo Zhao. 2021.
\newblock \href {https://doi.org/10.18653/v1/2021.acl-long.140} {Named entity
  recognition with small strongly labeled and large weakly labeled data}.
\newblock In \emph{Proceedings of the 59th Annual Meeting of the Association
  for Computational Linguistics and the 11th International Joint Conference on
  Natural Language Processing (Volume 1: Long Papers)}, pages 1775--1789,
  Online. Association for Computational Linguistics.

\bibitem[{Jiao et~al.(2020)Jiao, Yin, Shang, Jiang, Chen, Li, Wang, and
  Liu}]{jiao-etal-2020-tinybert}
Xiaoqi Jiao, Yichun Yin, Lifeng Shang, Xin Jiang, Xiao Chen, Linlin Li, Fang
  Wang, and Qun Liu. 2020.
\newblock \href {https://doi.org/10.18653/v1/2020.findings-emnlp.372}
  {{T}iny{BERT}: Distilling {BERT} for natural language understanding}.
\newblock In \emph{Findings of the Association for Computational Linguistics:
  EMNLP 2020}, pages 4163--4174, Online. Association for Computational
  Linguistics.

\bibitem[{Kaewphan et~al.(2016)Kaewphan, Landeghem, Ohta, de~Peer, Ginter, and
  Pyysalo}]{Kaewphan2016CellLN}
Suwisa Kaewphan, Sofie~Van Landeghem, Tomoko Ohta, Yves~Van de~Peer, Filip
  Ginter, and Sampo Pyysalo. 2016.
\newblock Cell line name recognition in support of the identification of
  synthetic lethality in cancer from text.
\newblock \emph{Bioinformatics}, 32:276 -- 282.

\bibitem[{Katiyar and Cardie(2018)}]{katiyar-cardie-2018-nested}
Arzoo Katiyar and Claire Cardie. 2018.
\newblock \href {https://doi.org/10.18653/v1/N18-1079} {Nested named entity
  recognition revisited}.
\newblock In \emph{Proceedings of the 2018 Conference of the North {A}merican
  Chapter of the Association for Computational Linguistics: Human Language
  Technologies, Volume 1 (Long Papers)}, pages 861--871, New Orleans,
  Louisiana. Association for Computational Linguistics.

\bibitem[{Kim and Kang(2022)}]{kim2022your}
Hyunjae Kim and Jaewoo Kang. 2022.
\newblock How do your biomedical named entity recognition models generalize to
  novel entities?
\newblock \emph{Ieee Access}, 10:31513--31523.

\bibitem[{Kim et~al.(2004)Kim, Ohta, Tsuruoka, Tateisi, and
  Collier}]{kim2004introduction}
Jin-Dong Kim, Tomoko Ohta, Yoshimasa Tsuruoka, Yuka Tateisi, and Nigel Collier.
  2004.
\newblock Introduction to the bio-entity recognition task at jnlpba.
\newblock In \emph{Proceedings of the international joint workshop on natural
  language processing in biomedicine and its applications}, pages 70--75.
  Association for Computational Linguistics.

\bibitem[{Krallinger et~al.(2015)Krallinger, Rabal, Leitner, Vazquez, Salgado,
  Lu, Leaman, Lu, Ji, Lowe et~al.}]{krallinger2015chemdner}
Martin Krallinger, Obdulia Rabal, Florian Leitner, Miguel Vazquez, David
  Salgado, Zhiyong Lu, Robert Leaman, Yanan Lu, Donghong Ji, Daniel~M Lowe,
  et~al. 2015.
\newblock The chemdner corpus of chemicals and drugs and its annotation
  principles.
\newblock \emph{Journal of cheminformatics}, 7.

\bibitem[{Lee et~al.(2020)Lee, Yoon, Kim, Kim, Kim, So, and
  Kang}]{lee2020biobert}
Jinhyuk Lee, Wonjin Yoon, Sungdong Kim, Donghyeon Kim, Sunkyu Kim, Chan~Ho So,
  and Jaewoo Kang. 2020.
\newblock Biobert: a pre-trained biomedical language representation model for
  biomedical text mining.
\newblock \emph{Bioinformatics}, 36(4):1234--1240.

\bibitem[{Lever et~al.(2020)Lever, Altman, and Kim}]{Lever2020ExtendingTF}
Jake Lever, Russ~B. Altman, and Jin-Dong Kim. 2020.
\newblock Extending textae for annotation of non-contiguous entities.
\newblock \emph{Genomics \& Informatics}, 18.

\bibitem[{Lewis et~al.(2020)Lewis, Ott, Du, and
  Stoyanov}]{lewis-etal-2020-pretrained}
Patrick Lewis, Myle Ott, Jingfei Du, and Veselin Stoyanov. 2020.
\newblock \href {https://doi.org/10.18653/v1/2020.clinicalnlp-1.17} {Pretrained
  language models for biomedical and clinical tasks: Understanding and
  extending the state-of-the-art}.
\newblock In \emph{Proceedings of the 3rd Clinical Natural Language Processing
  Workshop}, pages 146--157, Online. Association for Computational Linguistics.

\bibitem[{Li et~al.(2016)Li, Sun, Johnson, Sciaky, Wei, Leaman, Davis,
  Mattingly, Wiegers, and Lu}]{li2016biocreative}
Jiao Li, Yueping Sun, Robin~J Johnson, Daniela Sciaky, Chih-Hsuan Wei, Robert
  Leaman, Allan~Peter Davis, Carolyn~J Mattingly, Thomas~C Wiegers, and Zhiyong
  Lu. 2016.
\newblock Biocreative v cdr task corpus: a resource for chemical disease
  relation extraction.
\newblock \emph{Database: The Journal of Biological Databases \& Curation},
  2016.

\bibitem[{Liu et~al.(2020)Liu, Shareghi, Meng, Basaldella, and
  Collier}]{DBLP:journals/corr/abs-2010-11784}
Fangyu Liu, Ehsan Shareghi, Zaiqiao Meng, Marco Basaldella, and Nigel Collier.
  2020.
\newblock \href {http://arxiv.org/abs/2010.11784} {Self-alignment pre-training
  for biomedical entity representations}.
\newblock \emph{CoRR}, abs/2010.11784.

\bibitem[{Moritz et~al.(2018)Moritz, Nishihara, Wang, Tumanov, Liaw, Liang,
  Elibol, Yang, Paul, Jordan, and Stoica}]{ray-222605}
Philipp Moritz, Robert Nishihara, Stephanie Wang, Alexey Tumanov, Richard Liaw,
  Eric Liang, Melih Elibol, Zongheng Yang, William Paul, Michael~I. Jordan, and
  Ion Stoica. 2018.
\newblock \href {https://www.usenix.org/conference/osdi18/presentation/moritz}
  {Ray: A distributed framework for emerging {AI} applications}.
\newblock In \emph{13th USENIX Symposium on Operating Systems Design and
  Implementation (OSDI 18)}, pages 561--577, Carlsbad, CA. USENIX Association.

\bibitem[{Nakayama(2018)}]{seqeval}
Hiroki Nakayama. 2018.
\newblock \href {https://github.com/chakki-works/seqeval} {{seqeval}: A python
  framework for sequence labeling evaluation}.
\newblock Software available from https://github.com/chakki-works/seqeval.

\bibitem[{Neumann et~al.(2019)Neumann, King, Beltagy, and
  Ammar}]{neumann-etal-2019-scispacy}
Mark Neumann, Daniel King, Iz~Beltagy, and Waleed Ammar. 2019.
\newblock \href {https://doi.org/10.18653/v1/W19-5034} {{S}cispa{C}y: {F}ast
  and {R}obust {M}odels for {B}iomedical {N}atural {L}anguage {P}rocessing}.
\newblock In \emph{Proceedings of the 18th BioNLP Workshop and Shared Task},
  pages 319--327, Florence, Italy. Association for Computational Linguistics.

\bibitem[{Neves et~al.(2013)Neves, Damaschun, Mah, Lekschas, Seltmann,
  Stachelscheid, Fontaine, Kurtz, and Leser}]{neves2013preliminary}
Mariana Neves, Alexander Damaschun, Nancy Mah, Fritz Lekschas, Stefanie
  Seltmann, Harald Stachelscheid, Jean-Fred Fontaine, Andreas Kurtz, and Ulf
  Leser. 2013.
\newblock Preliminary evaluation of the cellfinder literature curation pipeline
  for gene expression in kidney cells and anatomical parts.
\newblock \emph{Database}, 2013.

\bibitem[{Ramshaw and Marcus(1999)}]{Ramshaw1999}
L.~A. Ramshaw and M.~P. Marcus. 1999.
\newblock \href {https://doi.org/10.1007/978-94-017-2390-9_10} {\emph{Text
  Chunking Using Transformation-Based Learning}}, pages 157--176. Springer
  Netherlands, Dordrecht.

\bibitem[{Ratner et~al.(2018)Ratner, Hancock, Dunnmon, Goldman, and
  R{\'e}}]{Ratner2018SnorkelMW}
Alexander~J. Ratner, Braden Hancock, Jared~A. Dunnmon, Roger~E. Goldman, and
  Christopher R{\'e}. 2018.
\newblock Snorkel metal: Weak supervision for multi-task learning.
\newblock \emph{Proceedings of the Second Workshop on Data Management for
  End-To-End Machine Learning}.

\bibitem[{Sennrich et~al.(2016)Sennrich, Haddow, and
  Birch}]{sennrich-etal-2016-neural}
Rico Sennrich, Barry Haddow, and Alexandra Birch. 2016.
\newblock \href {https://doi.org/10.18653/v1/P16-1162} {Neural machine
  translation of rare words with subword units}.
\newblock In \emph{Proceedings of the 54th Annual Meeting of the Association
  for Computational Linguistics (Volume 1: Long Papers)}, pages 1715--1725,
  Berlin, Germany. Association for Computational Linguistics.

\bibitem[{Smith et~al.(2008)Smith, Tanabe, nee Ando, Kuo, Chung, Hsu, Lin,
  Klinger, Friedrich, Ganchev et~al.}]{smith2008overview}
Larry Smith, Lorraine~K Tanabe, Rie~Johnson nee Ando, Cheng-Ju Kuo, I-Fang
  Chung, Chun-Nan Hsu, Yu-Shi Lin, Roman Klinger, Christoph~M Friedrich, Kuzman
  Ganchev, et~al. 2008.
\newblock Overview of biocreative ii gene mention recognition.
\newblock \emph{Genome biology}, 9(2):S2.

\bibitem[{Sung et~al.(2022)Sung, Jeong, Choi, Kim, Lee, and
  Kang}]{sung2022bern2}
Mujeen Sung, Minbyul Jeong, Yonghwa Choi, Donghyeon Kim, Jinhyuk Lee, and
  Jaewoo Kang. 2022.
\newblock Bern2: an advanced neural biomedical named entity recognition and
  normalization tool.
\newblock \emph{arXiv preprint arXiv:2201.02080}.

\bibitem[{Yoon et~al.(2019)Yoon, So, Lee, and Kang}]{yoon2019collabonet}
Wonjin Yoon, Chan~Ho So, Jinhyuk Lee, and Jaewoo Kang. 2019.
\newblock Collabonet: collaboration of deep neural networks for biomedical
  named entity recognition.
\newblock \emph{BMC bioinformatics}, 20(10):55--65.

\end{thebibliography}
\bibliographystyle{acl_natbib}

\appendix

%\section{Example Appendix}
%\label{sec:appendix}

%This is a section in the appendix.

\end{document}